\documentclass{article} 
\usepackage{iclr2025_conference,times}

\usepackage{amsmath,amsfonts,bm}

\def\eqref#1{equation~\ref{#1}}

\def\1{\bm{1}}

\DeclareMathAlphabet{\mathsfit}{\encodingdefault}{\sfdefault}{m}{sl}
\SetMathAlphabet{\mathsfit}{bold}{\encodingdefault}{\sfdefault}{bx}{n}

\usepackage{hyperref}
\usepackage{url}
\usepackage{booktabs}
\usepackage{amsfonts}
\usepackage{nicefrac}
\usepackage{microtype}  
\usepackage{xcolor} 
\usepackage{amsmath}
\usepackage{graphicx}
\usepackage{arydshln}
\usepackage{multirow}
\usepackage{bbding}
\usepackage{orcidlink}
\usepackage{amssymb}
\usepackage{wrapfig}

\usepackage{natbib}

\title{Hi-Mamba: Hierarchical Mamba for Efficient Image Super-Resolution}

\author{%
Junbo Qiao$^{1}$,\enspace Jincheng Liao$^{1}$,\enspace Wei Li$^{2}$,\enspace Yulun Zhang$^{3}$,\enspace Yong Guo$^{2}$,\enspace Yi Wen$^{2}$,\enspace \\ \textbf{Zhangxizi Qiu$^{2}$,\enspace Jiao Xie{$^{1}$},\enspace Jie Hu{$^{2}$},\enspace Shaohui Lin{$^{1}$}\thanks{Corresponding authors: Shaohui Lin, shlin@cs.ecnu.edu.cn}}\\
	\textsuperscript{1}East China Normal University,\enspace \textsuperscript{2}Huawei Technologies Ltd,\enspace \textsuperscript{3}Shanghai Jiaotong University
	\vspace{-4mm}
}

\iclrfinalcopy 
\begin{document}

\maketitle

\begin{abstract}
State Space Models (SSM), such as Mamba, have shown strong representation ability in modeling long-range dependency with linear complexity, achieving successful applications from high-level to low-level vision tasks. However, SSM's sequential nature necessitates multiple scans in different directions to compensate for the loss of spatial dependency when unfolding the image into a 1D sequence. This multi-direction scanning strategy significantly increases the computation overhead and is unbearable for high-resolution image processing. To address this problem, we propose a novel Hierarchical Mamba network, namely, Hi-Mamba, for image super-resolution (SR).  Hi-Mamba consists of two key designs: (1) The Hierarchical Mamba Block (HMB) assembled by a Local SSM (L-SSM) and a Region SSM (R-SSM) both with the single-direction scanning, aggregates multi-scale representations to enhance the context modeling ability. (2) The Direction Alternation  Hierarchical Mamba Group (DA-HMG) allocates the isomeric single-direction scanning into cascading HMBs to enrich the spatial relationship modeling.
Extensive experiments demonstrate the superiority of Hi-Mamba across five benchmark datasets for efficient SR. For example, Hi-Mamba achieves a significant PSNR improvement of 0.29 dB on Manga109 for $\times3$ SR, compared to the strong lightweight MambaIR.
\end{abstract}

\section{Introduction}
Single Image Super-Resolution~\cite{yang2019deep, he2019ode,zhang2018image,chen2022cross,zhang2021aligned} (SISR) aims to restore an authentic high-resolution (HR) image from a single degraded low-resolution (LR) one, which benefits plentiful downstream applications such as magnetic resonance imaging (MRI), mobile device photography, and video surveillance. Various studies have proposed Convolutional Neural Networks (CNNs)~\cite{ahn2018fast,li2021survey,zhang2019residual} to learn a mapping from LR inputs to HR outputs.
Despite their efficacy and remarkable advances in the past, CNN-based SR models are reaching their upper-performance limits even with continuously increasing model sizes, due to CNNs' limited capability on long-range dependency modeling.
%

Transformer-based SR methods~\cite{liang2021swinir,chen2023dual,chen2023activating,ray2024cfat,zhang2024transcending} introduce self-attention mechanisms with extraordinary long-range modeling capabilities to remarkably improve SR performance, while at the cost of quadratic computational complexity. 
%
%
Numerous subsequent works have been proposed to make the vanilla Transformers more efficient and powerful via shifted window attention~\cite{liang2021swinir,zhang2022efficient}, transposed attentions~\cite{zamir2022restormer,li2023dlgsanet} and anchored stripe self-attention~\cite{li2023efficient}, \emph{etc.} However, these studies are difficult to relieve the quadratic complexity of attention mechanisms at inference in practice.  

Recently, Mamba~\cite{gu2023mamba} architecture constructed on Structured State Space Models (S4) has emerged as a promising technique due to its high potential in long-sequence modeling with linear complexity.
As S4 was originally proposed in the field of natural language processing (NLP)~\cite{gu2021efficiently,gu2023mamba}, several succeeding works have introduced S4 into vision recognition tasks~\cite{liu2024vmamba,zhu2024vision} and image processing tasks~\cite{shi2024vmambair}, demonstrating impressive results.
For example, Vision Mamba~\cite{zhu2024vision} was proposed for image recognition tasks, manifesting that Vision Mamba can overcome the computation \& memory constraints on image perceptions. For low-level vision tasks, MambaIR~\cite{guo2024mambair} introduces the vision state-space module (VSSM) from Vmamba~\cite{liu2024vmamba} for image super-resolution and achieves performance comparable to Transformer-based SR baselines.


\begin{table}[t]
\small
\caption{Comparison of different scanning modes in MambaIR for $\times2$ SR. MambaIR-$n$ indicates using the number of $n$ scanning.}
\resizebox{1.0\columnwidth}{!}{
\begin{tabular}{c|ccc|ccccc}
\hline
    Method      & GPU(ms) & Params                     & FLOPs & Set5   & Set14 & B100  & Urban100 & Manga109 \\ \hline
MambaIR-1~\cite{guo2024mambair} & 519 & 987K & 291G          & 38.13  & 33.86 & 32.31 & 32.82    & 39.19    \\
MambaIR-2~\cite{guo2024mambair}  &653 & 1.11M &383G          &38.15 &33.94  &32.31 & 32.86 & 39.26    \\
MambaIR-4~\cite{guo2024mambair} &982 & 1.36M & 568G          & 38.16 & 34.00 & 32.34 & 32.92    & 39.31    \\ \hline
Hi-Mamba-S (Ours) & 379  & 1.34M & 274G          & 38.24 & 34.08 & 32.38 & 33.13    & 39.35    \\ \hline
\end{tabular}}
\label{tab:mambair}
\vspace{-.5cm}
\end{table}
Previous vision Mamba architectures typically employ a multi-direction scanning strategy to compensate for the loss of spatial dependencies when unfolding the image into a 1D sequence. 
Unfortunately, the repetitive multiple-sequence scanning overshadows the essential linear computational complexity of SSMs primarily designed with single-sequence scanning to model 1D sequential relationships. It significantly increases the computation overhead and is unacceptable for high-resolution image processing tasks.
%
As shown in Tab.~\ref{tab:mambair}, the four-sequence scanning approach effectively improves the performance by 0.10 dB and 0.12 dB on Urban100 and Manga109, respectively. However, this enhancement comes at a significant computational cost, increasing FLOPs by 95.2\% and parameters by 37.8\% compared to the single-sequence scanning approach in MambaIR.
%

%
To address this problem, we propose a novel Hierarchical Mamba architecture, termed Hi-Mamba, for image super-resolution (SR).  We first propose the hierarchical Mamba block (HMB) which is constructed by a local SSM and a region SSM with single-direction scanning to conduct multi-scale data-dependent visual context modeling. Furthermore, we propose the direction alternation hierarchical Mamba group (DA-HMG) that allocates the isomeric single-direction scanning into cascaded HMBs to enrich the spatial relationship modeling. Our DA-HMG improves the reconstruction performance with no extra FLOPs or parameter increases.
%
In addition, we propose that the gate feed-forward network (G-FFN) introduce additional non-linear information through a simple gate mechanism in the feed-forward network. 
We verify the effectiveness of Hi-Mamba on several classical SR benchmarks with three released versions, which makes fair comparisons with various SR models with different capacities.  

We summarize our main contributions as follows: 
\begin{itemize}
    \item 
    We propose Hi-Mamba for efficient SR, incorporating hierarchical Mamba block (HMB), specifically the Local-SSM and the Region-SSM for multi-scale data-dependent visual context modeling.
    \item 
    The direction alternation hierarchical Mamba group (DA-HMG) is simple yet effective in enriching the spatial relationship modeling, which allocates the isometric single-direction scanning into cascaded HMBs to improve performance without incurring extra computation and memory costs.
    \item 
    Extensive experiments demonstrate the superiority of the proposed Hi-Mamba. For example, our Hi-Mamba achieves significant PSNR gains of 0.37dB on Urban100 for $\times3$ SR compared to SRFormer~\cite{zhou2023srformer}.
    
\end{itemize}

\section{Related Work}
\subsection{Efficient CNNs and Transformers for Super-Resolution}
Since SRCNN\cite{dong2015image} first introduced convolutional neural networks (CNNs) for SR, various works~\cite{dong2016accelerating,lim2017enhanced,ledig2017photo,zhang2018image} have explored CNN-based SR architectures to improve SR performance. To improve model efficiency, CARN \cite{ahn2018fast} proposes a cascading mechanism at both the local and global levels. IMDN~\cite{hui2019lightweight} adopts feature splitting and concatenation operations to progressively aggregate features, further reducing parameters. SAFMN~\cite{sun2023safmn} utilizes a feature pyramid to generate spatially-adaptive feature attention maps. However, these CNN-based SR methods are limited by the size of the convolutional kernels and cannot effectively model long-term dependencies between pixels.

To capture long-range pixel dependencies, Transformer-based methods~\cite{liang2021swinir,chen2023dual,chen2023activating,ray2024cfat,zhang2024transcending} have introduced self-attention mechanisms into SR tasks, achieving significant performance improvements. To facilitate practical deployment, various efficient attention mechanisms~\cite{li2023efficient,zhou2023srformer} have been proposed to reduce computational and memory costs. ESRT~\cite{zhisheng2021efficient} computes attention maps in a group manner to reduce memory usage. N-Gram~\cite{choi2023n} proposed an asymmetric U-Net architecture that downsamples features to reduce computational cost. DLGSANet \cite{li2023dlgsanet} utilizes channel-wise self-attention, which has lower computational costs compared to spatial self-attention. SRFormer~\cite{zhou2023srformer} minimizes the size of the attention map by compressing the channel dimensions of the key and value in self-attention. However, these methods do not directly address the quadratically growing complexity of attention mechanisms with the increase in token sequence length. Moreover, they typically compute self-attention based on windows, which confines the receptive field for high-quality image reconstruction.

\begin{figure*}[t]
  \centering
   \includegraphics[width=0.98\linewidth]{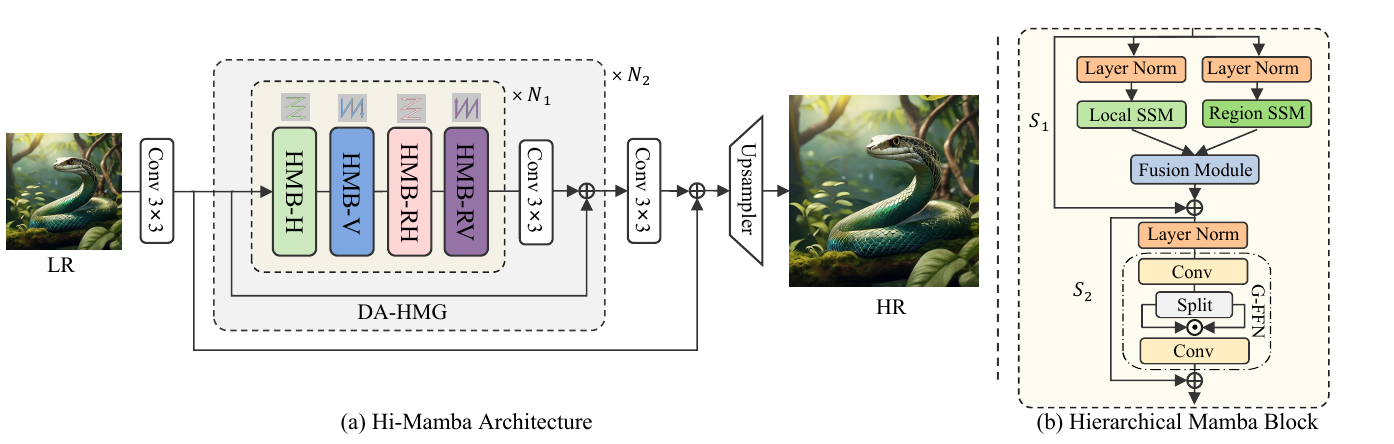}
   \caption{Illustration of the proposed Hi-Mamba. (a) The overview of Hi-Mamba architecture with $N_2$ Hierarchical Mamba Groups (DA-HMG), where each DA-HMG contains the number of $N_1$ Hierarchical Mamba blocks (HMB), which consist of four isomeric single-direction scanning SSM denoted by HMB-H/V/RH/RV. (b) Hierarchical Mamba Block (HMB) consists of a Local-SSM, a Region-SSM, and a Gate Feed-Forward Network (G-FFN).}
   \vspace{-.5cm}
   \label{fig:str}
   \vspace{-0.1cm}
\end{figure*}

\subsection{Mamba and Applications for Super-Resolution}
State space models (SSM)~\cite{gu2021efficiently,gu2021combining,smith2022simplified}, originating from classical control theory~\cite{kalman1960new}, are rising as novel backbones in Deep Learning. Successful applications of SSM include Mamba~\cite{gu2023mamba}, Vim~\cite{zhu2024vision}, and VMamba~\cite{liu2024vmamba}, which are all tailored toward high-level image understanding tasks. Overall, the implementations of SSM in low-level vision tasks remain few.
MambaIR~\cite{guo2024mambair} first introduced the Mamba architecture to image super-resolution tasks, achieving impressive image restoration results. MMA~\cite{cheng2024activating} introduced Vision Mamba (ViM)~\cite{zhu2024vision} and combined it with convolutional structures to activate a wider pixel area, thereby enhancing SR performance. DVMSR~\cite{lei2024dvmsr} was the first to attempt distilling the Mamba architecture to achieve an ultra-lightweight SR Mamba model. FMSR~\cite{xiao2024frequency} introduced Mamba for remote sensing image super-resolution, which uses frequency information to assist the Mamba architecture, achieving performance surpassing Transformer methods. However, these methods all use multi-sequence scanning strategies to model the image spatial relationships, which significantly increases computational costs compared to the single-sequence scanning of vanilla Mamba. 
Different from these methods, our Hi-Mamba uses only single-sequence scanning and proposes HMB to compensate for SSM's inadequacy in modeling 2D-pixel relationships. Additionally, the DA-HMG is proposed to enrich spatial relationship modeling by alternatively changing the single-sequence scanning direction in HMB without additional computational costs.


\section{Hierarchical Mamba Networks}
\label{method}

\subsection{Preliminaries}
SSM can be viewed as a Linear Time-Invariant (LTI) system, which maps the input one-dimensional function or sequence \( x(t) \in \mathbb{R} \) to the output response \( y(t) \in \mathbb{R} \) through a hidden state \( h(t) \in \mathbb{R}^N \). They are typically represented as linear ordinary differential equations:
\begin{equation}
\label{eq1}
h^{\prime}(t)=A h(t)+ B x(t), \quad
y(t)=C h(t)+D x(t),
\end{equation}
where \( A \in \mathbb{R}^{N \times N} \), \( B \in \mathbb{R}^{N \times 1} \), \( C \in \mathbb{R}^{1 \times N} \), and \( D \in \mathbb{R} \) are weight parameters, and \( N \) represents the state size.

The discretization process is commonly used to process Eq.~\ref{eq1}, which can be applied in deep learning scenarios. 
In particular, the timescale parameter \( \Delta \) is used to convert the continuous parameters \( A \) and \( B \) into discrete ones $\overline{A}$ and $\overline{B}$. The widely used discretization method adheres to the Zero-Order Hold (ZOH) rule, which is formulated as:
\begin{equation}
\label{eq2}
 \overline{A}=\exp (\Delta A), \quad
\overline{B}=(\Delta A)^{-1}(\exp (A)-I) \cdot \Delta B.
\end{equation}
Therefore, after discretization, Eq.~\ref{eq1} can be rewritten as:
\begin{equation}
\label{eq3}
h_k =\overline{A} h_{k-1}+\overline{B} x_k, \quad
y_k =C h_k+D x_k.
\end{equation}
To further accelerate computation, Gu et al.~\cite{gu2021efficiently} expanded the SSM computation into a convolution with a structured convolutional kernel $\overline{K}\in \mathbb{R}^L$:
\begin{equation}
\label{eq4}
\overline{K} \triangleq\left(C \overline{B}, C \overline{A B}, \cdots, C \overline{A}^{L-1} \overline{B}\right), \quad
y  =x *  \overline{K},
\end{equation}
where \( L \) is the length of the input sequence and \( * \) denotes the convolution operation. A recent state space model, Mamba~\cite{gu2023mamba}, introduces Selective State Space Models (S6) by relaxing the time-invariance constraints on \( B \), \( C \), and \( \Delta \) depending on the input $x$, which selectively propagates information for 1D language sequence modeling.

To expand Mamba from 1D language sequences to 2D visual inputs, various works~\cite{liu2024vmamba,liang2024pointmamba,deng2024cu,guo2024mambair} employ 2D selective scan (SS2D) mechanism to capture spatial correlations with 2D feature sequences. For example, VMamba~\cite{liu2024vmamba} employs SS2D by scanning four directed input sequences and generating the 2D feature map by independently combining four feature sequences via an S6 block. Similarly, MambaIR~\cite{guo2024mambair} introduces the Vision State-Space Module (VSSM) into image restoration for information interaction at the whole-image level. However, these methods employ repetitive multi-direction scanning to adapt to 2D image inputs, significantly increasing computational costs.

\subsection{Architecture Overview}
As shown in Fig.~\ref{fig:str} (a), the proposed Hierarchical Mamba (Hi-Mamba) architecture comprises three parts: shallow feature extraction, deep feature extraction, and image reconstruction. Given a low-resolution (LR) input image \( I_{\text{LR}} \in \mathbb{R}^{C_{\text{in}} \times H \times W} \), where \( C_{\text{in}} \), \( H \), and \( W \) are the input channels, height, and width, respectively. We first use a simple convolution for shallow feature extraction $H_{SF}$ to generate local features $F_{l} \in \mathbb{R}^{C \times H \times W }$:
\begin{equation}
F_{l}=H_{SF}(I_{LR}),
\end{equation}
where $C$ is the embedding channel dimension. Subsequently, the local features $F_{l}$ are processed in the deep feature extraction module $H_{DF}$ to obtain deep features $F_{d} \in \mathbb{R}^{C \times H \times W }$:
\begin{equation}
F_{d}=H_{DF}(F_{l}),
\end{equation}
where the deep feature extraction module $H_{DF}$ consists of multiple direction alternation hierarchical Mamba groups (DA-HMG) with a total number of $N_2$. To ensure training stability, a residual strategy is adopted within each group. 
Each DA-HMG contains the number of $N_1$ Hierarchical Mamba blocks (HMB), which consist of four isomeric single-direction scanning SSM denoted by HMB-H/V/RH/RV. 
At the end of each DA-HMG, convolutional layers are introduced to refine the features. 

Finally, we use $F_{l}$ and $F_{d}$ as the inputs and reconstruct the high-resolution (HR) output image $H_{R}$ through the reconstruction module, which can be formulated as:
\begin{equation}
I_{r}=H_{R}(F_{l}+F_{d}),
\end{equation}
where $H_{R}$ involves a single 3$\times$3 convolution followed by a pixel shuffle operation. We optimize the parameters $\theta$ of Hi-Mamba by the pixel-wise L1 loss between the reconstruction output $I_{r}$ and the ground truth (GT) $I_{gt}$. In the following, we will introduce the key blocks and modules in Hi-Mamba.

\begin{figure*}[t]
  \centering
   \includegraphics[width=0.9\linewidth]{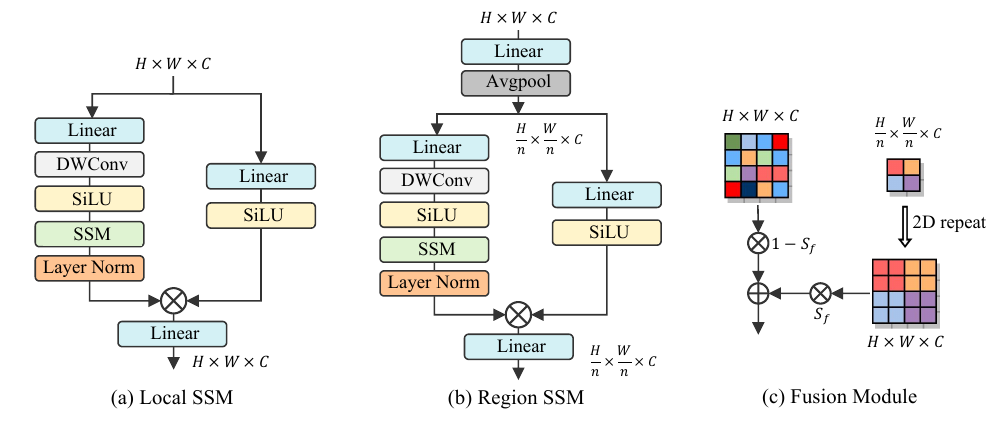}
   \caption{Illustration of the key components in HMB.}
   \vspace{-.5cm}
   \label{fig:local&region}
\end{figure*}

\subsection{Hierarchical Mamba Block}
\label{hmb}
The original visual Mamba blocks~\cite{liu2024vmamba,zhu2024vision} typically employ multi-direction scanning, which significantly increases the computation overhead. To address this problem, we design a novel hierarchical mamba block (HMB) with only single-direction scanning and alternatively change the scanning direction to enrich the spatial relationship modeling to construct DA-HMG.


As illustrated in Fig.~\ref{fig:str} (b), HMB primarily consists of two branches: Local SSM (L-SSM) and Region SSM (R-SSM). 
Given the local input feature $I^i_l \in \mathbb{R}^{C \times H \times W }$ and the region input feature $I^i_r \in \mathbb{R}^{C \times \frac{H}{n} \times \frac{W}{n} }$ at the $i$-th layer, we first employ Layer Normalization (LN) and go through two branches to capture long-range dependencies.
Additionally, we incorporate learnable scaling factors \( S_1 \in \mathbb{R}^C \) to regulate the information within skip connections:
\begin{equation}
\begin{gathered}
    F^i_l = \operatorname{L-SSM}(\operatorname{LN}(I^i_l)),  F^i_r = \operatorname{R-SSM}(\operatorname{LN}(I^i_r)), \\
    F^i = (F^i_l \otimes F^i_r) + (S_1 \cdot I^i_l).
\end{gathered}
\end{equation}
where $F^i_l$, and $F^i_r$ are the outputs of these two branches, respectively. $\otimes$ denotes the fusion module. The L-/R-SSM and fusion modules will be described in Sec.~\ref{L-SSM} and Sec.~\ref{FM}, respectively.

Subsequently, the intermediate features $F^i$ will subsequently undergo the proposed gate feed-forward network (G-FFN) followed by another learnable scale factor $S_2$ in the residual connection to obtain the input features at the $i+1$-th layer and $F^i_r$ is directly used as the regional input for the next layer:
\begin{equation}
F^{i+1}_l=\operatorname{G-FFN}\left(\operatorname{LN}(F^i)\right)+S_2 \cdot F^i, \quad F^{i+1}_r = F^i_r.
\end{equation}

In G-FFN, we enhance the modeling capacity for spatial information by introducing a gate mechanism into the FFN. This also reduces redundant information in the channels. G-FFN first extracts features through convolution and splits the feature map along the channel dimension into two parts for element-wise multiplication. Specifically, G-FFN is computed as:
\begin{equation}
\begin{aligned}
& \hat{F}=w^1*\operatorname{LN}(F^i), [\hat{F}_1, \hat{F}_2]=\text{Split}(\hat{F}), \\
& \operatorname{G-FFN}(F^i)=w^2*(\hat{F}_1 \odot \hat{F}_2),
\end{aligned}
\end{equation}
where $w^1$ and $w^2$ are the convolution weights. $\odot$ is an element-wise multiplication operation.
Note that, we only use a single-direction scanning in one HMB, \emph{i.e.}, one selection from the horizontal, vertical, reverse horizontal, and reverse vertical directions.
\subsubsection{Local~/~Region SSM}
\label{L-SSM}
Following the VSSM of MambaIR~\cite{guo2024mambair}, L-SSM and R-SSM use a similar computational sequence. 
Instead of VSSM with multiple-sequence scanning, L-SSM and R-SSM employ single-sequence scanning to reduce the computation costs.
The architecture of L-SSM and R-SSM are illustrated in Fig.~\ref{fig:local&region} (a) and (b), respectively.
L-SSM and R-SSM take the local feature $I_l \in \mathbb{R}^{C \times H \times W}$ and the region feature $I_r \in \mathbb{R}^{C \times \frac{H}{n} \times \frac{W}{n}}$ as the inputs, respectively. Here, $I_r$ is generated by a simple projection operation with a region size of $n$ to the local feature $I_l$.
%
For simplicity, we denote the input uniformly as $X$, due to the same computation process to L-SSM and R-SSM. 

In the first branch of L-SSM, feature channels are expanded to \( \lambda C \) via a linear layer, where \( \lambda \) is a predefined channel expansion factor, followed by depthwise convolution, SiLU~\cite{shazeer2020glu} activation function, SSM and LayerNorm. In the second branch, feature channels are also expanded to \( \lambda C \) with a linear layer and SiLU activation function. Finally, the features from both branches are merged and projected back to \( C \) to generate an output \( X_{\text{out}} \) with the same shape as the input. The above computation process can be formulated as:
\begin{equation}
\label{local_region}
\begin{split}
& X_{b1} = \operatorname{LN}(\mathrm{SSM}(\operatorname{SiLU}(\mathrm{DWConv}(\operatorname{Linear}(X))))), \\
& X_{b2} = \operatorname{SiLU}(\operatorname{Linear}(X)), \\
& X_{out}= \operatorname{Linear}(X_{b1} \odot X_{b2}),
\end{split}
\end{equation}
where DWConv($\cdot$), SSM($\cdot$) and \( \odot \) represent depthwise convolution, SS2D~\cite{liu2024vmamba} with single-direction scanning and element-wise multiplication, respectively.
\subsubsection{Fusion Module}
\label{FM}
To reinforce spatial dependencies in the 2D domains, we use the fusion module to leverage region information from adjacent pixels in the R-SSM to guide the single-sequence local feature modeling. As illustrated in Fig.~\ref{fig:local&region} (c), we first repeat the region features along the spatial dimension to match the size of the local features, ensuring that each region token is mapped to the corresponding local token. 
This operation implicitly incorporates spatial positional information. 
To dynamic control the fusion results, we introduce learnable fusion scaling factors $S_f \in \mathbb{R}^C$ to fuse the outputs of L-SSM and R-SSM in Eq.~\ref{local_region}, which is formulated as:
\begin{equation}
F_{out}=S_f \cdot X^l_{out} + (1-S_f) \cdot f_{re}(X^r_{out}).
\end{equation}
where $X^l_{out}$ and $X^r_{out}$ denote the outputs of the L-SSM and R-SSM, respectively. $f_{re}$ represents the repeat operation along the 2D spatial dimension.

\subsection{Direction Alternation Hierarchical Mamba Group}
\label{HMG}
As depicted in Fig.~\ref{fig:str}, DA-HMG is easy to implement by alternatively allocating the isomeric single-direction scanning to different HMBs. By default, we apply Horizontal HMB (HMB-H), Vertical HMB (HMB-V), Reverse Horizontal HMB (HMB-RH), and Reverse Vertical HMB (HMB-RV) orders to enrich the spatial relationship modeling further. DA-HMG does not incur extra parameters and computational costs, compared to the HMB with the same direction, denoted by base-single. 

Compared to the stacked multi-sequence scanning in the 2D-SSM module of MambaIR, DA-HMG significantly reduces the computational and parameter overhead while achieving superior performance. The more detailed difference between base-single, 2D-SSM and DA-HMG on the sequence scanning strategy is presented in Fig. \ref{fig:ssm_scan} of Appendix.


%
%
%

\section{Experiments}
\subsection{Experimental Settings}
\label{sec:detail}
\textbf{Datasets.} 
Following~\cite{liang2021swinir,guo2024mambair,li2023feature,chen2023dual}, we train our model on two widely-used datasets, DIV2K~\cite{agustsson2017ntire} and Flicker2K~\cite{lim2017enhanced}, and only use DIV2K dataset to train the lightweight version of our model. We evaluate our method on five standard SR benchmarks: Set5~\cite{bevilacqua2012low}, Set14~\cite{zeyde2012single}, BSD100~\cite{martin2001database}, Urban100~\cite{huang2015single}, and Manga109~\cite{matsui2017sketch} across three scaling factors, $\times 2$, $\times 3$, and $\times 4$. For the evaluation metric, we calculate PSNR and SSIM~\cite{wang2004image} on the Y channel in the YCbCr space and also report the average inference time (20 runs) on one NVIDIA V100, parameters and FLOPs.

\textbf{Implementation details.} 
Following the general setting~\cite{liang2021swinir,zhang2022accurate,chen2022cross}, each training sample is augmented through flipping and rotations of $90^{\circ}$, $180^{\circ}$ and $270^{\circ}$. During training, we randomly crop images into $64 \times 64$ patches, with a total iteration number of 500K. The patch size is set to 32. We employ the Adam optimizer with training parameters $\beta_1 = 0.9$, $\beta_2 = 0.999$, and zero weight decay. The initial learning rate was 2e-4, which was halved at iterations [250K, 400K, 450K, 475K]. The experiments are implemented by PyTorch using 8 NVIDIA V100 GPUs.
We provide three versions of Hi-Mamba with varying complexities, denoted as Hi-Mamba-T, Hi-Mamba-S and Hi-Mamba-L. The details of the three versions can be found in the Appendix. 

\begin{table}[t]
	\centering
	\scriptsize       
	\caption{Quantative comparison of lightweight SR models on five benchmarks. The {\color[HTML]{FF0000}best} and {\color[HTML]{0000FF}second-best} results for Transformers and Mamba are marked in red and blue colors. }
        \vspace{.5em}
        \resizebox{.99\columnwidth}{!}{
        \label{tab:light}
	\begin{tabular}{|c|c|c|c|cc|cc|cc|cc|cc|}
		\hline
		\multirow{2}*{\textbf{Scale}}        & \multirow{2}*{\textbf{Model}}              & \multicolumn{1}{c|}{\textbf{Params}} & \multicolumn{1}{c|}{\textbf{FLOPs}} & \multicolumn{2}{c|}{\textbf{Set5}}                       & \multicolumn{2}{c|}{\textbf{Set14}}                      & \multicolumn{2}{c|}{\textbf{BSD100}}                       & \multicolumn{2}{c|}{\textbf{Urban100}}                       & \multicolumn{2}{c|}{\textbf{Manga109}} \\
  &  & (M)& (G) & PSNR & SSIM & PSNR & SSIM & PSNR & SSIM & PSNR & SSIM &PSNR & SSIM \\ \hline
  \multirow{13}*{x2}  & CARN~\cite{ahn2018fast}   & 1.45 & 223& 37.76& 0.9590 & 33.52 & 0.9166 & 32.09 & 0.8978 & 31.92 & 0.9256 & 38.36 & 0.9765 \\
    & EDSR-baseline~\cite{lim2017enhanced}   &1.37 & 316 & 37.99 &0.9604 &33.57 &0.9175 & 32.16 &0.8994 & 31.98 &0.9272 & 38.54 &0.9769\\
    & IMDN~\cite{hui2019lightweight}   & 0.69  & 159   & 38.00  & 0.9605 & 33.63  & 0.9177 & 32.19   & 0.8996  & 32.17   & 0.9283  & 38.88  & 0.9774  \\
    & LAPAR-A~\cite{li2020lapar}  & 0.55 & 171 & 38.01 &0.9605  & 33.62&0.9183 & 32.19&0.8999 & 32.10&0.9283& 38.67&0.9772 \\
    & LatticeNet~\cite{luo2020latticenet}   & 0.76  & 170  & 38.15  & 0.9610 & 33.78  & 0.9193 & 32.25  & 0.9005  & 32.43  & 0.9302  & -  & - \\
    \cdashline{2-14}[1pt/1pt]
    
    & ESRT~\cite{zhisheng2021efficient} & 0.67 &  - & 38.03 &0.9600& 33.75 & 0.9184& 32.25 &0.9001 & 32.58 &0.9318& 39.12 & 0.9774\\
    & SwinIR-Light~\cite{liang2021swinir}  & 0.90 & 235  & 38.14&0.9611& 33.86&0.9206&  32.31&0.9012 & 32.76&0.9340& 39.12&0.9783\\
    & N-Gram~\cite{choi2023n} & 1.01 & 140 & 38.05 & 0.9610& 33.79 & 0.9199& 32.27 & 0.9008& 32.53&0.9324 & 38.97&0.9777\\
    & SRFormer-Light~\cite{zhou2023srformer}   & 0.83 & 236  & {\color[HTML]{0000FF}38.23}   & {\color[HTML]{0000FF}0.9613}  & 33.94 & 0.9209 & {\color[HTML]{0000FF}32.36} & {\color[HTML]{0000FF}0.9019} & 32.91 & 0.9353 & 39.28  & 0.9785 \\
    \cdashline{2-14}[1pt/1pt]
    &MambaIR~\cite{guo2024mambair}  &1.36& 568  &38.16&0.9610 &34.00&0.9212  &32.34&0.9017 &32.92&0.9356 &{\color[HTML]{0000FF}39.31}&0.9779 \\
    &Hi-Mamba-T  & 0.87 & 178  &{\color[HTML]{FF0000}38.24}&{\color[HTML]{0000FF}0.9613} &{\color[HTML]{0000FF}34.06}&{\color[HTML]{0000FF}0.9215} &32.35&{\color[HTML]{0000FF}0.9019}&{\color[HTML]{0000FF}33.04}&{\color[HTML]{0000FF}0.9358}& 39.28&{\color[HTML]{0000FF}0.9785}\\
   &Hi-Mamba-S  & 1.34 & 274  &{\color[HTML]{FF0000}38.24}&{\color[HTML]{FF0000}0.9614} &{\color[HTML]{FF0000}34.08}&{\color[HTML]{FF0000}0.9217} &{\color[HTML]{FF0000}32.38}&{\color[HTML]{FF0000}0.9021}&{\color[HTML]{FF0000}33.13}&{\color[HTML]{FF0000}0.9368}& {\color[HTML]{FF0000}39.35}&{\color[HTML]{FF0000}0.9788}\\
    \hline 
    \multirow{13}*{x3}& CARN~\cite{ahn2018fast}  & 1.59  & 119  & 34.29&0.9255& 30.29&0.8407& 29.06&0.8034& 28.06&0.8493& 33.50&0.9440\\
    & EDSR-baseline~\cite{lim2017enhanced} & 1.56  & 160  & 34.37&0.9270& 30.28&0.8417& 29.09&0.8052& 28.15&0.8527& 33.45&0.9439\\
    & IMDN~\cite{hui2019lightweight}   &0.70  &72   &34.36  &0.9270 &30.32   &0.8417 &29.09   &0.8046  &28.17   &0.8519   &33.61   &0.9445  \\
    & LAPAR-A~\cite{li2020lapar}  & 0.54  & 114 & 34.36&0.9267& 30.34&0.8421& 29.11&0.8054& 28.15&0.8523& 33.51&0.9441\\
    & LatticeNet~\cite{luo2020latticenet}   &0.77  &76   &34.53  &0.9281 &30.39   &0.8424 &29.15   &0.8059  &28.33   & 0.8538  & -  & - \\
    \cdashline{2-14}[1pt/1pt]

    & ESRT~\cite{zhisheng2021efficient}  & 0.77  & -  & 34.42&0.9268& 30.43&0.8433& 29.15&0.8063& 28.46&0.8574& 33.95&0.9455\\
    & SwinIR-Light~\cite{liang2021swinir} & 0.89 & 87 & 34.62&0.9289& 30.54&0.8463& 29.20&0.8082 &28.66&0.8624& 33.98&0.9478\\
    & N-Gram~\cite{choi2023n}& 1.01 & 67 & 34.52&0.9282& 30.53&0.8456& 29.19&0.8078& 28.52&0.8603& 33.89&0.9470\\
    & SRFormer-Light~\cite{zhou2023srformer}  &0.86  &105   &34.67  &0.9296 &30.57  &0.8469 &29.26 &0.8099 &28.81 &0.8655  &34.19  &0.9489 \\
    \cdashline{2-14}[1pt/1pt]
    &MambaIR~\cite{guo2024mambair} & 1.37 & 253 & 34.72&0.9296 & {\color[HTML]{0000FF}30.63}&{\color[HTML]{0000FF}0.8475} & {\color[HTML]{0000FF}29.29}&0.8099& 29.00&0.8689 &34.39&0.9495 \\
    &Hi-Mamba-T  & 0.88& 80  & {\color[HTML]{0000FF}34.76}&{\color[HTML]{0000FF}0.9298} &30.61&0.8472 &29.27&{\color[HTML]{0000FF}0.8091} &{\color[HTML]{0000FF}29.05}&{\color[HTML]{0000FF}0.8693} &{\color[HTML]{0000FF}34.42}&{\color[HTML]{0000FF}0.9499} \\
    &Hi-Mamba-S  & 1.35& 123  & {\color[HTML]{FF0000}34.77}&{\color[HTML]{FF0000}0.9303} &{\color[HTML]{FF0000}30.68}&{\color[HTML]{FF0000}0.8493} &{\color[HTML]{FF0000}29.33}&{\color[HTML]{FF0000}0.8111} &{\color[HTML]{FF0000}29.18}&{\color[HTML]{FF0000}0.8716} &{\color[HTML]{FF0000}34.68}&{\color[HTML]{FF0000}0.9509} \\
    \hline

    \multirow{13}*{x4}& CARN~\cite{ahn2018fast}  & 1.59 & 91 & 32.13&0.8937& 28.6&0.7806 & 27.58&0.7349& 26.07&0.7837& 30.47&0.9084\\
    & EDSR-baseline~\cite{lim2017enhanced}& 1.52 & 114 & 32.09&0.8938& 28.58&0.7813& 27.57&0.7357& 26.04&0.7849& 30.35&0.9067\\
    & IMDN~\cite{hui2019lightweight}   &0.72  &41   &32.21  &0.8948 &28.58   &0.7811 &27.56   &0.7353  &26.04   &0.7838   &30.45   &0.9075  \\
    & LAPAR-A~\cite{li2020lapar}  & 0.66  & 94 & 32.15&0.8944& 28.61&0.7818& 27.61&0.7366& 26.14&0.7871& 30.42&0.9074\\
    & LatticeNet~\cite{luo2020latticenet}  &0.78  &44   &32.30  &0.8962 &28.68   &0.7830 & 27.62  &0.7367  &26.25   &0.7873   &  -  &  - \\
    \cdashline{2-14}[1pt/1pt] 
    & ESRT~\cite{zhisheng2021efficient} & 0.75  &  64& 32.19&0.8947& 28.69&0.7833& 27.69&0.7379& 26.39&0.7962& 30.75&0.9100\\
    & SwinIR-Light~\cite{liang2021swinir} & 0.90 & 50 & 32.44&0.8976& 28.77&0.7858& 27.69&0.7406& 26.47&0.7980& 30.92&0.9151\\
    & N-Gram~\cite{choi2023n}& 1.02 & 36 & 32.33&0.8963& 28.78&0.7859& 27.66&0.7396& 26.45&0.7963& 30.80&0.9128\\
    & SRFormer-Light~\cite{zhou2023srformer}  &0.87  &63   &32.51  &0.8988 &{\color[HTML]{0000FF}28.82}  &0.7872 &27.73 &0.7422 &26.67 &0.8032  &31.17  &0.9165 \\
    \cdashline{2-14}[1pt/1pt]
    &MambaIR~\cite{guo2024mambair} &1.40 & 143 &32.51&0.8993 & {\color[HTML]{0000FF}28.82}&{\color[HTML]{0000FF}0.7876} & 27.65&0.7423 & 26.75&0.8051 & 31.26&0.9175\\
    &Hi-Mamba-T & 0.89 & 45 & {\color[HTML]{0000FF}32.52}&{\color[HTML]{0000FF}0.8995} &28.80&0.7873  &{\color[HTML]{0000FF}27.75}&{\color[HTML]{0000FF}0.7429 }&{\color[HTML]{0000FF}26.81}&{\color[HTML]{0000FF}0.8072 }&{\color[HTML]{0000FF}31.35}&{\color[HTML]{0000FF}0.9186} \\
    &Hi-Mamba-S & 1.36 & 69 & {\color[HTML]{FF0000}32.60}&{\color[HTML]{FF0000}0.8999} &{\color[HTML]{FF0000}28.91}&{\color[HTML]{FF0000}0.7895 } &{\color[HTML]{FF0000}27.78}&{\color[HTML]{FF0000}0.7436 }&{\color[HTML]{FF0000}26.86}&{\color[HTML]{FF0000}0.8086 }&{\color[HTML]{FF0000}31.46}&{\color[HTML]{FF0000}0.9192 }\\
    
    \hline  
    	\end{tabular}}
    \end{table}

\begin{figure*}[t]
	\centering
	\begin{minipage}{0.33\linewidth}
		\vspace{4pt}
		\centerline{\includegraphics[width=\textwidth]{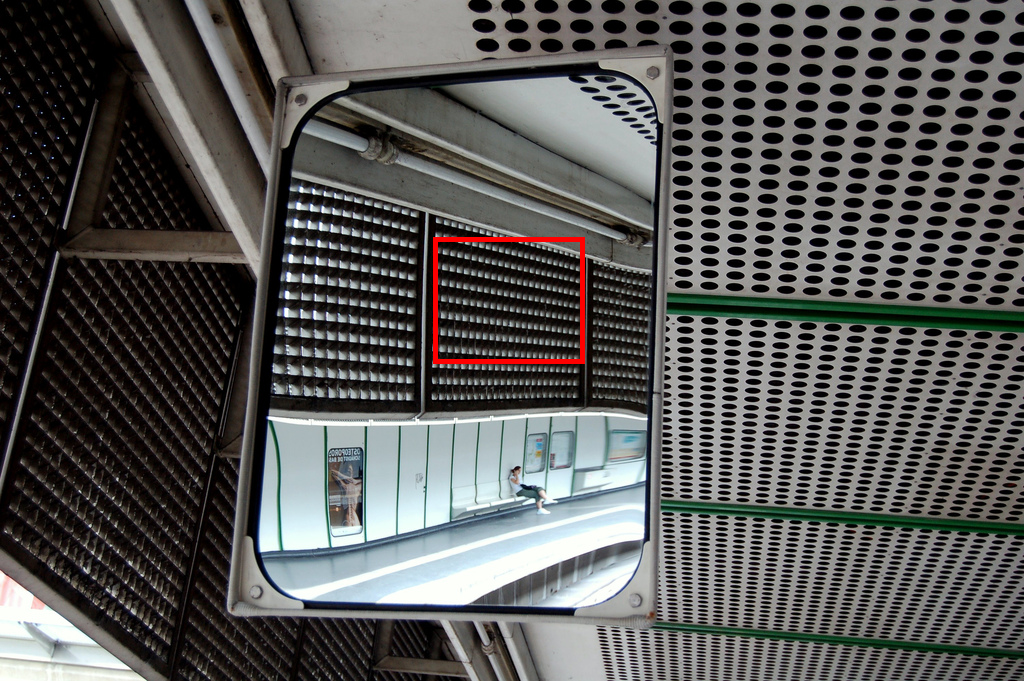}}
		\centerline{\scriptsize img004 from Urban100}
		\vspace{3pt}

	\end{minipage}
	\begin{minipage}{0.11\linewidth}
		\centerline{\includegraphics[width=\textwidth]{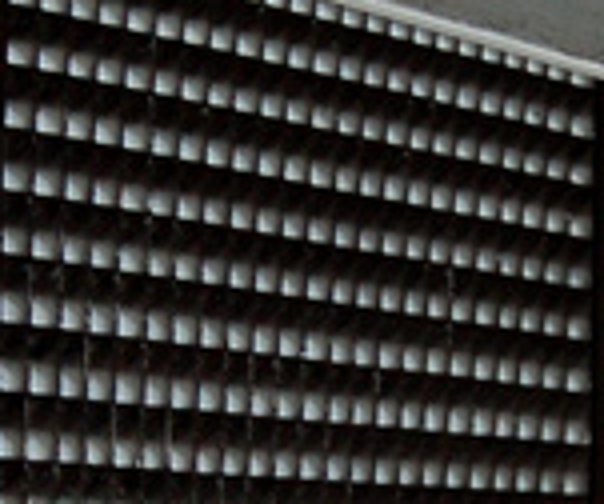}}
		\centerline{\tiny (a)HR}
		\centerline{\includegraphics[width=\textwidth]{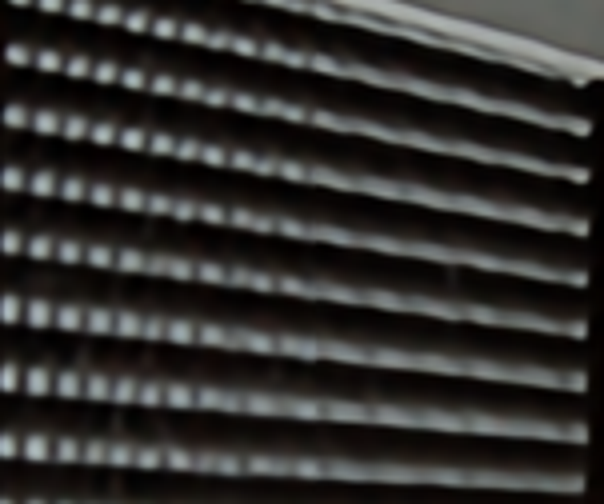}}
		\centerline{\tiny (f)ESRT } 

	\end{minipage}
	\begin{minipage}{0.11\linewidth}
		\centerline{\includegraphics[width=\textwidth]{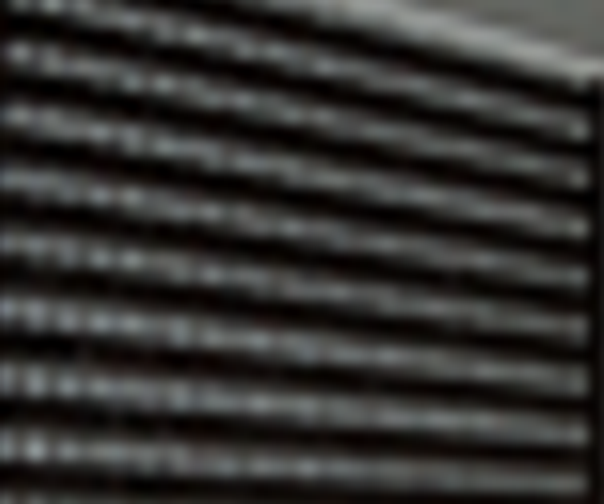}}
		\centerline{\tiny (b)Bicubic}
		\centerline{\includegraphics[width=\textwidth]{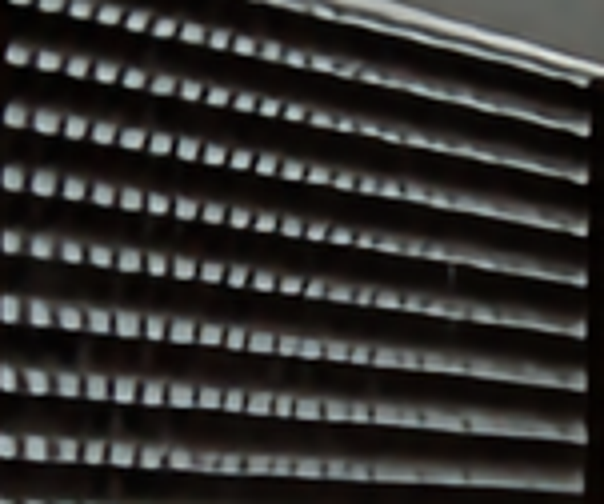}}
		\centerline{\tiny (g)SwinIR-light} 

	\end{minipage}
	\begin{minipage}{0.11\linewidth}
		\centerline{\includegraphics[width=\textwidth]{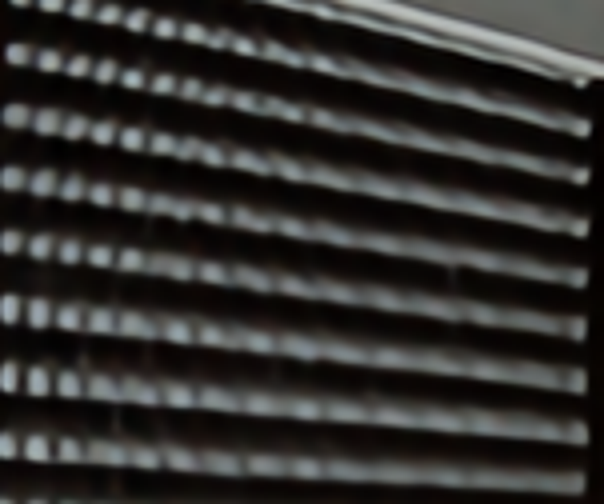}}
		\centerline{\tiny (c)CARN} 

		\centerline{\includegraphics[width=\textwidth]{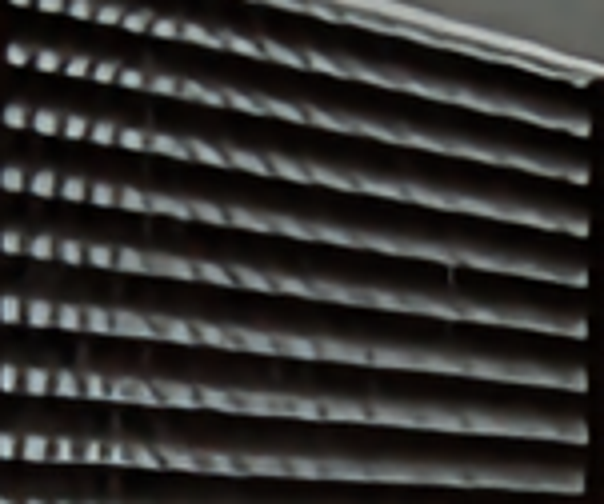}}
		\centerline{\tiny (h)N-Gram} 

	\end{minipage}
	\begin{minipage}{0.11\linewidth}
		\centerline{\includegraphics[width=\textwidth]{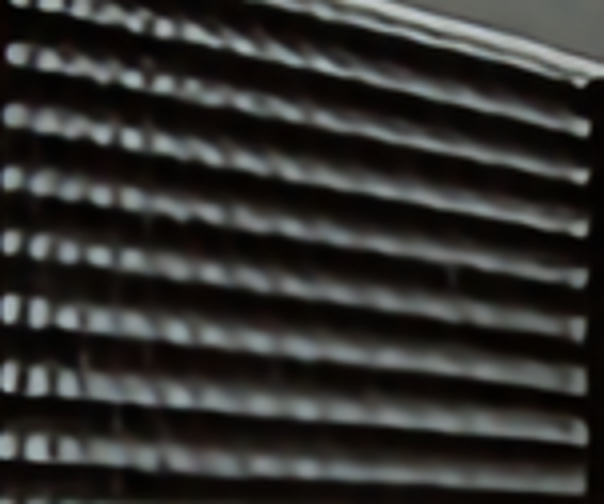}}
		\centerline{\tiny (d)EDSR } 
		\centerline{\includegraphics[width=\textwidth]{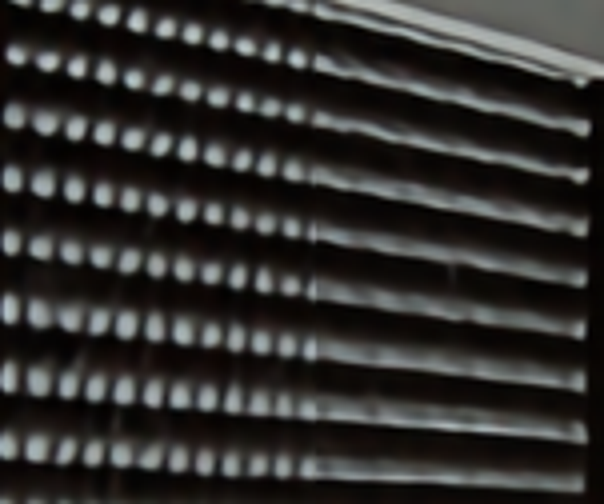}}
		\centerline{\tiny (i)MambaIR} 

	\end{minipage}
	\begin{minipage}{0.11\linewidth}
		\centerline{\includegraphics[width=\textwidth]{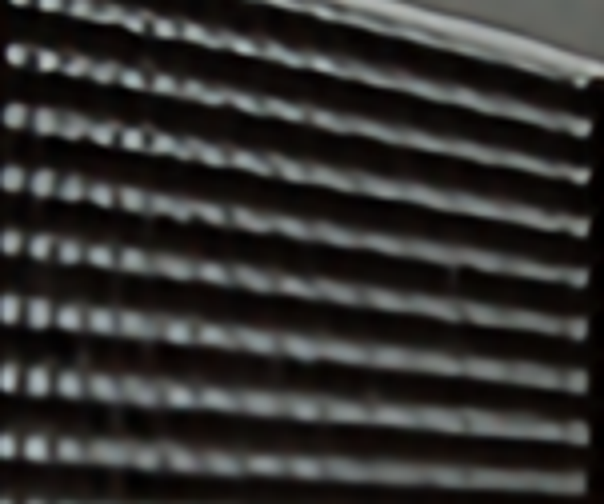}}
		\centerline{\tiny (e)IMDN} 
		\centerline{\includegraphics[width=\textwidth]{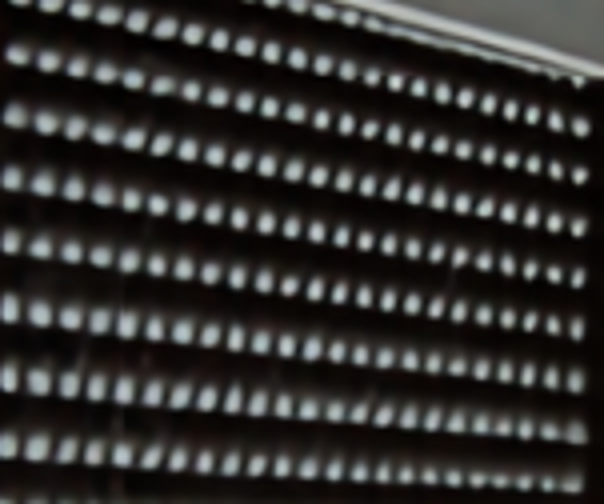}}
		\centerline{\tiny (j)Hi-Mamba-S}

	\end{minipage}
        \caption{Qualitative comparison on the ``img004'' image of Urban100 for $\times$4 SR.}
         \label{fig:x4_light_visual}
	\vspace{-.5cm}
\end{figure*}

\subsection{Comparison with Lightweight SR models.}

\textbf{Quantitative evaluations.} Tab.~\ref{tab:light} summarizes the quantitative results at three SR scale factors of $\times 2$, $\times 3$ and $\times 4$. The parameter and computational costs of MambaIR\cite{guo2024mambair} are modified by the tool\footnote{{\href{https://github.com/MzeroMiko/VMamba/blob/main/classification/models/vmamba.py\#L1372}{https://github.com/MzeroMiko/VMamba/blob/main/classification/models/vmamba.py\#L1372}}}
Compared to CNN-based methods, Transformer-based approaches (such as IMDN \cite{hui2019lightweight} and SRFormer-Light \cite{zhou2023srformer}) introduce self-attention mechanisms to model long-range dependencies, exhibiting superior performance in terms of PSNR and SSIM. Notably, transformer-based methods often utilize window-based self-attention mechanisms in the super-resolution task to reduce computational but limit the receptive field within the window. In contrast, MambaIR employs the SSM to model long-range dependencies, which outperforms the SOTA SRFormer~\cite{zhou2023srformer} by 0.09 PSNR on Urban100 for 3$\times$ SR. However, MambaIR requires $1.59 \times$ parameter and $2.41 \times$ FLOPs compared to SRFormer. This is due to the usage of computation-heavy multi-sequence directional scanning in SSM and the redundant structural design.
%
%
%
For a fair comparison, we compare the proposed Hi-Mamba-T and Hi-Mamba-S with state-of-the-art lightweight SR methods.
Benefiting from the multi-scale mechanism and DA-HMG, Hi-Mamba-T and Hi-Mamba-S outperform SRFormer and MambaIR in terms of PSNR and SSIM across multiple benchmark datasets with comparable parameters and FLOPs. For example, compared to MambaIR, Hi-Mamba-S and Hi-Mamba-T reduce FLOPs by 294G and 390G, while improving the PSNR for $\times 2$ SR on Urban100 by 0.21 dB and 0.12 dB, respectively. Meanwhile, Hi-Mamba-T significantly outperforms SRFormer by 0.24 dB on $\times 3$ scale SR on Urban100, while reducing 25 GFLOPs and maintaining relatively consistent parameters.
%

\begin{figure*}
    \centering
    \includegraphics[width=0.8\linewidth]{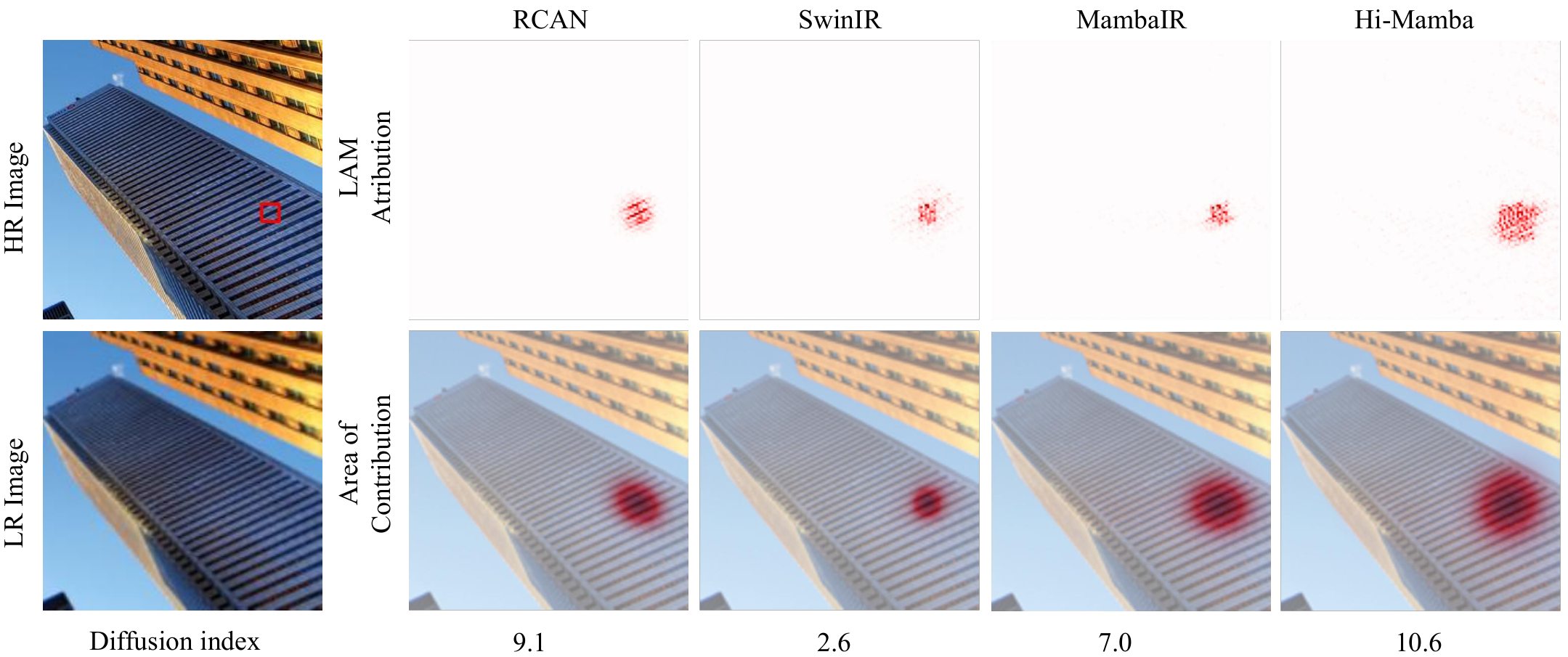}
    \vspace{-.1cm}
    \caption{LAM visualization~\cite{gu2021interpreting} on $\times2$ SR task. LAM indicates the correlation between the significance of each pixel in LR and the SR patch outlined with the red box. Hi-Mamba utilizes a broader range of information to obtain better performance.}
    \label{fig:lam}
    \vspace{-.6cm}
\end{figure*}

\begin{table*}[t]
	\centering
	\scriptsize
	\caption{Comparison of different PSNR-oriented SR models on five benchmarks. Methods with ”*” are replicated with standard setting, detailed in the Appendix. Methods with "+" denote the use of a self-ensemble strategy.}
    \resizebox{.99\columnwidth}{!}{
	\label{tab:base}
	\begin{tabular}{|c|c|cc|cc|cc|cc|cc|}
		\hline
		\multirow{2}*{\textbf{Model}}        & \multirow{2}*{\textbf{Scale}} & \multicolumn{2}{c|}{\textbf{Set5}}                       & \multicolumn{2}{c|}{\textbf{Set14}}                      & \multicolumn{2}{c|}{\textbf{BSD100}}                       & \multicolumn{2}{c|}{\textbf{Urban100}}                       & \multicolumn{2}{c|}{\textbf{Manga109}} \\
   &   &PSNR &SSIM &PSNR &SSIM &PSNR &SSIM &PSNR &SSIM &PSNR &SSIM \\ \hline
   
		EDSR~\cite{lim2017enhanced} & $\times 2$ & 38.11&0.9602 & 33.92&0.9195 & 32.32&0.9013 & 32.93&0.9351 & 39.10&0.9773    \\
		RCAN~\cite{zhang2018image} & $\times 2$ & 38.27&0.9614 & 34.12&0.9216 &32.41&0.9027 & 33.34&0.9384 & 39.44&0.9786  \\
            SAN~\cite{dai2019second} & $\times 2$ & 38.31&0.9620 & 34.07&0.9213 & 32.42&0.9028 & 33.10&0.9370  & 39.32&0.9792 \\
		HAN~\cite{niu2020single} & $\times 2$ & 38.27&0.9614 & 34.16&0.9217 & 32.41&0.9027 & 33.35&0.9385 & 39.46&0.9785 \\
            IGNN~\cite{zhou2020cross} & $\times 2$ &38.24 &0.9613 &34.07 &0.9217 &32.41 &0.9025 &33.23 &0.9383 &39.35 &0.9786 \\
            CSNLN~\cite{mei2020image} & $\times 2$ &38.28 &0.9616 &34.12 &0.9223 &32.40 &0.9024 &33.25 &0.9386 &39.37 &0.9785 \\
		NLSN~\cite{mei2021image} & $\times 2$ & 38.34&0.9618 & 34.08&0.9231 &32.43&0.9027  & 33.42&0.9394  & 39.59&0.9789 \\ \cdashline{1-12}[1pt/1pt] 
            ELAN~\cite{zhang2022efficient} & $\times 2$  & 38.36&0.9620 &33.20&0.9228 & 32.45&0.9030 & 33.44&0.9391 & 39.62&0.9793 \\
            DLGSANet~\cite{li2023dlgsanet} & $\times 2$ & 38.34&0.9617 &34.25&0.9231 & 32.38&0.9025 & 33.41&0.9393 & 39.57&0.9789 \\
            IPT~\cite{chen2021pre} & $\times 2$ &38.37 & - &34.43 &- &32.48 & - &33.76 &- &- &- \\
		SwinIR~\cite{liang2021swinir} & $\times 2$ & 38.42&0.9623 & 34.46&0.9250 
            &32.53&0.9041 & 33.81&0.9427 & 39.92&0.9797  \\
            EDT~\cite{li2021efficient} & $\times 2$ & 38.45&0.9624 & 34.57&0.9258 & 32.52&0.9041 & 33.80&0.9425 & 39.93&0.9800 \\
            GRL-B*~\cite{li2023efficient} & $\times 2$ & 38.48&0.9627 &34.64&0.9265 & 32.55&0.9045 &33.97&0.9437 & 40.06&0.9804 \\
		SRFormer~\cite{zhou2023srformer} & $\times 2$ & 38.51&0.9627 &34.44&0.9253 & 32.57&0.9046 & 34.09&0.9449 & 40.07&0.9802 \\
            
            \cdashline{1-12}[1pt/1pt] 
            MambaIR~\cite{guo2024mambair} & $\times 2$ &38.57&0.9627 & 34.67&0.9261 &32.58&0.9048 & 34.15&0.9466 & 40.28&0.9806 \\ 
		\textbf{Hi-Mamba-L}      & $\times 2$    & {\color[HTML]{0000FF}38.58}  &{\color[HTML]{0000FF}0.9633} &{\color[HTML]{0000FF}34.70} &{\color[HTML]{0000FF}0.9264} {\color[HTML]{0000FF} } & {\color[HTML]{0000FF} 32.60} &{\color[HTML]{0000FF}0.9054} {\color[HTML]{0000FF}} & {\color[HTML]{0000FF} 34.22} & {\color[HTML]{0000FF} 0.9475} &{\color[HTML]{0000FF}40.38} &{\color[HTML]{0000FF}0.9820} \\ 
        \textbf{Hi-Mamba-L+}      & $\times 2$      &{\color[HTML]{FF0000}38.60} &{\color[HTML]{FF0000}0.9634} &{\color[HTML]{FF0000}34.78} &{\color[HTML]{FF0000}0.9269} &{\color[HTML]{FF0000}32.63} &{\color[HTML]{FF0000}0.9058} &{\color[HTML]{FF0000}34.34}  &{\color[HTML]{FF0000}0.9483} &{\color[HTML]{FF0000}40.49} &{\color[HTML]{FF0000}0.9822} \\\hline
  EDSR~\cite{lim2017enhanced} & $\times 4$ & 32.46&0.8968 & 28.80&0.7876 & 27.71&0.7420 & 26.64&0.8033 & 31.02&0.9148    \\
		RCAN~\cite{zhang2018image} & $\times 4$ & 32.63&0.9002 & 28.87&0.7889 &27.77&0.7436 & 26.82&0.8087 & 31.22&0.9173  \\
            SAN~\cite{dai2019second} & $\times 4$ & 32.64&0.9003 & 28.92&0.7888 & 27.78&0.7436 & 26.79&0.8068  & 31.18&0.9169 \\
		HAN~\cite{niu2020single} & $\times 4$ & 32.64&0.9002 & 28.90&0.7890 & 27.80&0.7442 & 26.85&0.8094 & 31.42&0.9177 \\
            IGNN~\cite{zhou2020cross} & $\times 4$ &32.57 &0.8998 &28.85 &0.7891 &27.77 &0.7434 &26.84 &0.8090 &31.28 &0.9182 \\
            CSNLN~\cite{mei2020image} & $\times 4$ &32.68 &0.9004 &28.95 &0.7888 &27.80 &0.7439 &27.22 &0.8168 &31.43 &0.9201 \\
		NLSN~\cite{mei2021image} & $\times 4$ & 32.59&0.9000 & 28.87&0.7891 &27.78&0.7444  & 26.96&0.8109  & 31.27&0.9184 \\ \cdashline{1-12}[1pt/1pt] 
            ELAN~\cite{zhang2022efficient} & $\times 4$  & 32.75&0.9022 &28.96&0.7914 & 27.83&0.7459 & 27.13&0.8167 & 31.68&0.9226 \\
            DLGSANet~\cite{li2023dlgsanet} & $\times 4$ & 32.80&0.9021 &28.95&0.7907 & 27.85&0.7464 & 27.17&0.8175 & 31.68&0.9219 \\
            IPT~\cite{chen2021pre} & $\times 4$ &32.64 &- &29.01 &- &27.82 &- &27.26 &- &- &- \\
		SwinIR~\cite{liang2021swinir} & $\times 4$ & 32.92&0.9044 & 29.09&0.7950 
            & 27.92&0.7489 & 27.45&0.8254 & 32.03&0.9260  \\
            EDT~\cite{li2021efficient} & $\times 4$ & 32.82&0.9031 & 29.09&0.7939 & 27.91&0.7483 & 27.46&0.8246 & 32.05&0.9254 \\
            GRL-B*~\cite{li2023efficient} & $\times 4$ & 32.90&0.9039 &29.14&0.7956 & 27.96&0.7497 & 27.53&0.8276 & 32.19&0.9266 \\
		SRFormer~\cite{zhou2023srformer} & $\times 4$ & 32.93&0.9041 &29.08&0.7953 & 27.94&0.7502 & 27.68&{\color[HTML]{FF0000}0.8311} & 32.21&0.9271 \\
            \cdashline{1-12}[1pt/1pt] 
            MambaIR~\cite{guo2024mambair} & $\times 4$ & 33.03&0.9046 & 29.20&0.7961 & 27.98&0.7503 & 27.68&0.8287 & 32.32&0.9272 \\ 
        \textbf{Hi-Mamba-L}      & $\times 4$      & {\color[HTML]{0000FF}33.05}&{\color[HTML]{0000FF}0.9049} &{\color[HTML]{0000FF}29.23} &{\color[HTML]{0000FF}0.7966} & {\color[HTML]{0000FF}28.01} &{\color[HTML]{0000FF}0.7531} & {\color[HTML]{0000FF} 27.72} &0.8296 &{\color[HTML]{0000FF}32.43} &{\color[HTML]{0000FF}0.9280} \\ 
        \textbf{Hi-Mamba-L+}      & $\times 4$     & {\color[HTML]{FF0000}33.08}&{\color[HTML]{FF0000}0.9051} &{\color[HTML]{FF0000}29.26} &{\color[HTML]{FF0000}0.7969} & {\color[HTML]{FF0000}28.02} &{\color[HTML]{FF0000}0.7534} & {\color[HTML]{FF0000} 27.81} & {\color[HTML]{0000FF} 0.8304} &{\color[HTML]{FF0000}32.56} &{\color[HTML]{FF0000}0.9300} \\ \hline
    	\end{tabular}}
 \vspace{-2em}
\end{table*}

\textbf{Qualitative Comparison.} 
In Fig.~\ref{fig:x4_light_visual}, we present the visual comparisons for $\times4$ SR. We can observe that previous CNN-based or Transformer-based methods suffer from blurry artifacts, distortions, and inaccurate texture restoration. 
In contrast, our method effectively reduces these artifacts, preserving more structural and clear details. More visual examples can be referred to in the Appendix. 
Moreover, as shown in Fig.\ref{fig:lam}, we also visualize the Local Attribution Map~(LAM)~\cite{gu2021interpreting} to demonstrate the strong ability for long-range modeling using our Hi-Mamba-S.



\subsection{Comparison with PSNR-oriented SR Models}

To validate the scalability of Hi-Mamba, we further compare our Hi-Mamba-L with state-of-the-art PSNR-oriented SR models. 

\textbf{Quantitative evaluations.} Tab.~\ref{tab:base} summarizes the SR results at the scales of $\times2$ and $\times4$. 

\begin{wrapfigure}{r}{6cm}
  \centering
   \includegraphics[width=1.\linewidth]{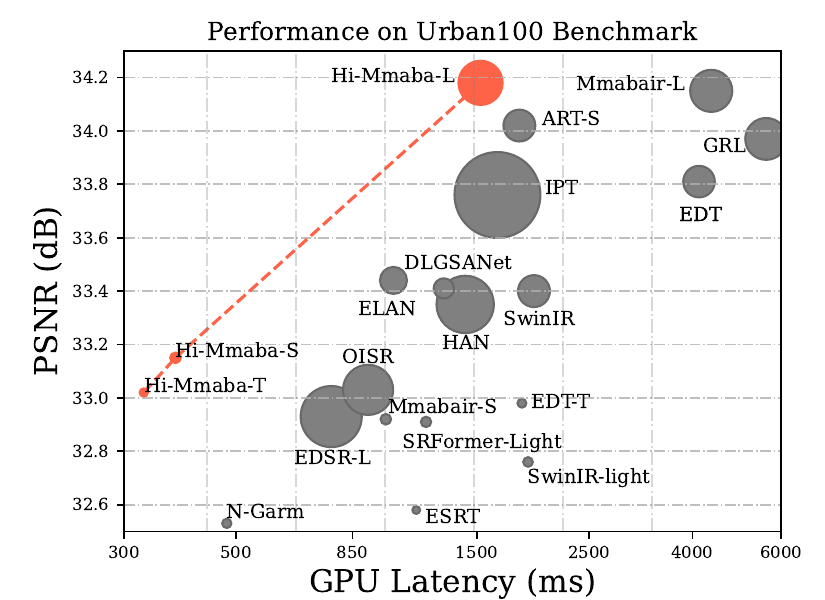}
   \vspace{-1em}
   \caption{Performance on Urban100 for $\times 2$ SR. The larger circles present larger computation costs on Params.}
   \vspace{-2em}
   \label{fig:psnr-latency}
\end{wrapfigure}

Our Hi-Mamba-L demonstrates superior performance compared to previous methods. In addition, the performance of Hi-Mamba-L can be further improved by using the self-ensemble strategy, denoted by Hi-Mamba-L+. For example, compared to SRFormer~\cite{zhou2023srformer} and MambaIR~\cite{guo2024mambair}, our Hi-Mamba-L+ achieves significant PSNR gains of 0.42 dB and 0.21 dB on Manga109 for $\times2$ SR, respectively.
For $\times4$ SR, our Hi-Mamba-L+ outperforms SRFormer by the PSNR of 0.18 dB and 0.35 dB on Set14 and Manga109, respectively. $\times 3$ SR result is presented in the Appendix.

\begin{figure*}[t]
	\centering
	\begin{minipage}{0.35\linewidth}
		\vspace{4pt}
		\centerline{\includegraphics[width=\textwidth]{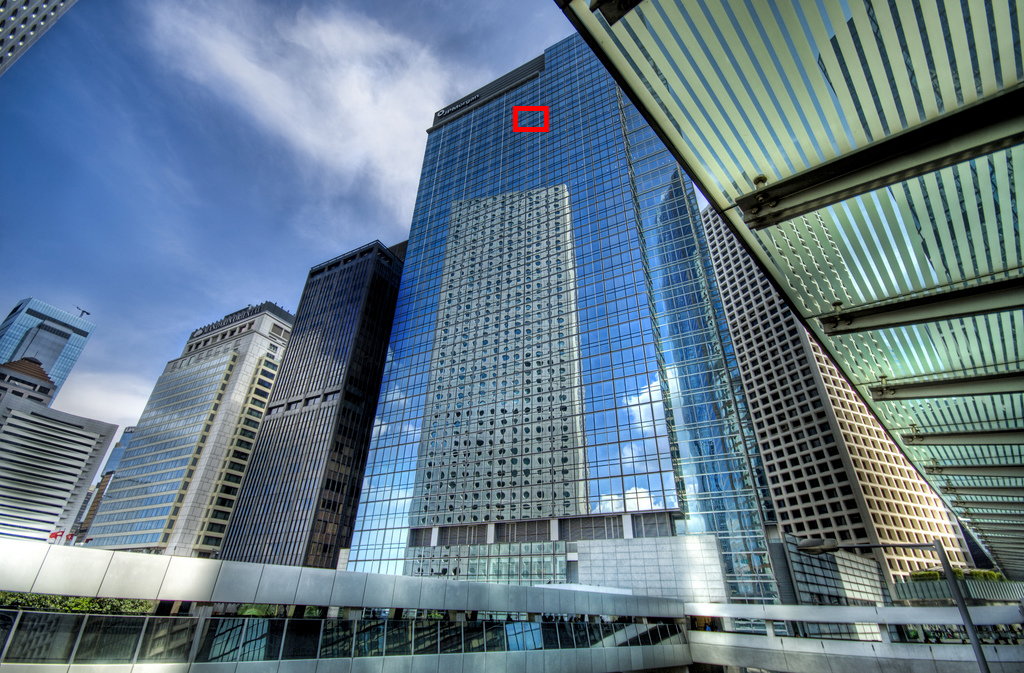}}
		\centerline{\scriptsize img061 from Urban100}
		\vspace{3pt}

	\end{minipage}
	\begin{minipage}{0.14\linewidth}
		\centerline{\includegraphics[width=\textwidth]{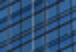}}
		\centerline{\tiny (a)HR}
		\centerline{\includegraphics[width=\textwidth]{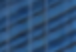}}
		\centerline{\tiny (e) DLGSANet} 

	\end{minipage}
	\begin{minipage}{0.14\linewidth}
		\centerline{\includegraphics[width=\textwidth]{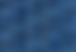}}
		\centerline{\tiny (b)Bicubic}
		\centerline{\includegraphics[width=\textwidth]{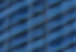}}
		\centerline{\tiny (f)SRFormer} 

	\end{minipage}
	\begin{minipage}{0.14\linewidth}
		\centerline{\includegraphics[width=\textwidth]{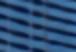}}
		\centerline{\tiny (c)RCAN} 
		\centerline{\includegraphics[width=\textwidth]{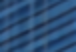}}
		\centerline{\tiny (g)MambaIR} 

	\end{minipage}
	\begin{minipage}{0.14\linewidth}
		\centerline{\includegraphics[width=\textwidth]{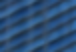}}
		\centerline{\tiny (d)SwinIR} 
		\centerline{\includegraphics[width=\textwidth]{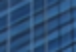}}
		\centerline{\tiny (h)Hi-Mamba-L}

	\end{minipage}
	\vspace{-1em}
        \caption{Qualitative comparison on the ``img061'' image of Urban100 for $\times$4 SR.}
         \label{fig:x4_classical_visual_main}
	\vspace{-0.6cm}
\end{figure*}

\textbf{Qualitative comparison.} 
We present the visual comparison of classic SR (×4) in Fig.~\ref{fig:x4_classical_visual_main}. Compared with CNNs (e.g., RCAN) and Transformers (e.g., SwinIR, SRFormer, DLGSANet), as well as SSM-based MambaIR, Hi-Mamba reconstructs the most photo-realistic building texture compared to these models.

\textbf{Model Complexity Comparison.} 
Tab.~\ref{tab:class_flops} further makes our Hi-Mamba-L with CNNs and Transformers in terms of parameters and FLOPs. Our Hi-Mamba-L significantly reduces GFLOPs by 7,702 and 1,881, and achieves 0.22dB and 0.10dB PSNR gains on Manga109 over GRL-B and MambaIR. To evaluate the practical inference time, we conduct the experiments on the PSNR and speed results of different methods as shown in Fig.~\ref{fig:psnr-latency}. We can observe that Hi-Mamba achieves the best latency-PSNR trade-off. 


\begin{table}[]
\centering
\scriptsize
\caption{Model complexity comparisons ($\times 2$). PSNR (dB) on Urban100 and Manga109, FLOPs, and Params are reported. Methods with "*" are replicated with standard settings.}
\begin{tabular}{l|ccccccccc}
\hline
\multirow{1}{*}{Method} & EDSR & RCAN & SAN & HAN & NLSA & SwinIR &GRL-B* & MambaIR & Hi-Mamba-L \\ \hline
Params(M)               &43.09      &15.59      & 15.87    & 63.61    &41.80      & 11.90  &20.20       &20.42         & 21.58         \\
FLOPs(G)                &11,130      & 3,530     &3,050     & 14,551    &9,632      & 3,213  &12,036     &  6,215       & 4,334         \\
PSNR-Urban100(dB)                &32.93      &33.34      & 33.10    &33.35     &33.42      &33.81   & 33.97    & 34.15        &34.22          \\
PSNR-Managa109(dB)               & 39.10     &  39.44    & 39.32    & 39.46    & 39.59     &  39.92  & 40.06   &  40.28       &   40.38       \\ \hline
\end{tabular}
\label{tab:class_flops}
\vspace{-0.7cm}
\end{table}

\subsection{Ablation Studies}
In the ablation study, we train the models on DIV2K evaluated on Urban100 for $2\times$ SR, as it contains images with rich structural details. For a fair comparison, we train the \emph{baseline} composed of only L-SSM and MLP stacks with a depth number equal to Hi-Mamba-S.

\textbf{Ablation for key components of Hi-Mamba.} We first conduct the ablation study on the effect of R-SSM, G-FFN, and DA-HMG. As shown in Tab.~\ref{tab:main}, the R-SSM significantly improves PSNR by 0.19 dB. With the FFN replaced by G-FFN, this model achieves a gain of 0.04 dB over baseline+R-SMM while reducing 0.1M parameters and 15G FLOPs. Finally, by utilizing DA-HMG, we further improve PSNR by 0.14 dB without incurring additional computational costs. This indicates that all the key components of Hi-Mamba show their effectiveness.

\begin{minipage}{\textwidth}

\begin{minipage}[t]{0.55\textwidth}
\makeatletter\def\@captype{table}
\centering 
\footnotesize
\caption{Ablation study of the key components. }
\resizebox{1.0\columnwidth}{!}{
\begin{tabular}{ccccccc}
\hline
R-SSM & G-FFN & DA-HMG & Params(M) & FLOPs(G) & PSNR  & SSIM \\ \hline
         &      &     & 1.29      &   252       & 32.76 & 0.9339     \\
       \Checkmark  &      &     & 1.44      &     289     & 32.95 &0.9354      \\
        \Checkmark & \Checkmark&     & 1.34       &   274       & 32.99 & 0.9356     \\
       \Checkmark &  \Checkmark& \Checkmark& 1.34       &   274       & 33.13 & 0.9368     \\ \hline
\end{tabular}}
\label{tab:main}

\centering 
\footnotesize
\caption{Ablation study of DA-HMG.}
\resizebox{.95\columnwidth}{!}{
\begin{tabular}{cccccc}
\hline
Model & PSNR(dB) & SSIM  &Params(M) &FLOPs(G)  \\ \hline
Single-direction w/o alternation  &32.99& 0.9356 &1.34 &274 \\
Two-direction alternation  &33.07&0.9361&1.34 &274 \\
Four-direction alternation  &33.13&0.9368 &1.34 &274\\ \hline
\end{tabular}}
\label{tab:scan}

\end{minipage}
\hspace{0.2cm}
\begin{minipage}[t]{0.45\textwidth}
\makeatletter\def\@captype{table}

\centering 
\footnotesize
\caption{Effect of fusion module in R-SSM.} 
\resizebox{1.\columnwidth}{!}{
\begin{tabular}{cccc}
\hline
Fusion method & Upsampling & Repeat & Repeat $S_f=0.5$ \\ \hline
Params(M) &1.34 &1.34 &1.33 \\
FLOPs(G)  &275 &274 &274\\
GPU(ms)  & 387 &379 &371\\
PSNR(dB)  &33.06& 33.13 &33.08 \\
SSIM  &0.9352&0.9368 &0.9360\\ \hline
\end{tabular}}
\label{tab:fusion}

\centering 
\footnotesize
\caption{Abaliton of R-SSM channel number.}
\resizebox{.70\columnwidth}{!}{
\begin{tabular}{c|ccc}
\hline
\#Channel & FLOPs(G) & Params(M) & PSNR(dB) \\ \hline
15        & 252      & 1.30      & 33.01    \\
30        & 274      & 1.34       & 33.13    \\
60        & 296      & 1.76      & 33.14    \\ \hline
\end{tabular}
\label{tab:R-SSM_d}
}

\end{minipage}
\end{minipage}

\begin{table}[h]
\centering 
\footnotesize
\vspace{-0.5cm}
\caption{Effect of region size in R-SSM. } 
\begin{tabular}{cccccc}
\hline
Region size & Params(M) & FLOPS(G) & GPU(ms) & PSNR(dB) & SSIM   \\ \hline
$1 \times 1$         & 1.52      & 312      & 662     & 33.13    & 0.9369 \\
$4 \times 4$          & 1.34      & 274      & 379     & 33.13    & 0.9368 \\
$8 \times 8$          & 1.34      & 271      & 365     & 33.01    & 0.9358 \\ \hline
\end{tabular}
\label{tab:region_size}
\end{table}

\textbf{Ablation for different scan modes in DA-HMG.} 
To investigate the effect of different scan modes in DA-HMG, we compare four-direction alternative scanning with single-direction without alternation(i.e., base-single), and two-direction alternative scanning, as summarized in Tab.~\ref{tab:scan}. By default, single-direction without alternation only uses HMB-H, and two-direction alternative scanning uses HMB-H and HMB-V.
%
We observe that the model using four-direction alternation achieves the best performance with an improvement of 0.14 dB PSNR and 0.06 dB PSNR over single-direction without alternation and two-direction alternation, respectively. Note that alternative direction scanning does not incur additional computational and memory costs. This indicates that direction alternation in DA-HMG can aggregate spatial information from different positions to improve reconstruction performance.

\textbf{Effect of fusion module in R-SSM.} As shown in Tab.~\ref{tab:fusion}, the repeat method implicitly incorporates the 2D spatial position information of features, achieving a PSNR of 0.07 dB higher than the upsampling method. It demonstrates the effectiveness of our 2D repeat fusion method. We also conduct additional ablation experiments on the learnable parameters $S_f$. 
We observed that the learnable parameter $S_f$ achieves only a slight increase of 0.01M parameters and 8ms GPU while outperforming fixed $S_f$ by 0.05dB PSNR. Thus, we default use the learnable $S_f$ for the fusion module.

\textbf{Ablation of R-SSM channel number.} We analyze the computational complexity of hierarchical design to achieve the best PSNR-FLOPs trade-off by changing the channel number of R-SSM. 
In Tab.~\ref{tab:R-SSM_d}, the channel number of 30 in R-SSM (\emph{i.e.}, a half of L-SSM) achieves the best trade-off between performance and computation complexity.

\textbf{Ablation for the region size $n\times n$ in R-SSM.} As presented in Tab.~\ref{tab:region_size}, we find that a region patch size of 1$\times$1 achieves the highest SSIM of 0.9369, but the inference time significantly increases, compared to patch sizes of 4$\times$4 and 8$\times$8. The region size of $4 \times 4$ yields the best trade-off between PSNR and inference speed. Thus, we set the region size to $4 \times 4$ for our experiments.

\section{Conclusion}
We present the Hierarchical Mamba Network (Hi-Mamba) in this paper for image super-resolution.  Hi-Mamba is built on multiple-direction alternation hierarchical Mamba groups (DA-HMG), which allocates the isomeric single-direction scanning into cascading HMBs, enriching the modeling of spatial relationships. Each HMB consists of a Local SSM and a Region SSM, utilizing unidirectional scanning to aggregate multi-scale representations and enhance 2D spatial perception.  Extensive experiments demonstrate that our Hi-Mamba has high potential compared to CNN-based and transformer-based methods.

\section*{Reproducibility Statement}
In this section, we provide a reproducibility statement for our proposed method. We detail the model architecture and core designs in Sec.~\ref{method}, including the hierarchical mamba block (HMB) and Direction Alternation HMB Group (DA-HMG). Additionally, we present implementation details and elaborate on the experimental setup in Sec.~\ref{sec:detail}. To ensure reproducibility, we will release the source code and pre-trained models. For more details, please refer to the Appendix.

\bibliography{arxiv}
\bibliographystyle{iclr2025_conference}

\end{document}